# Skin lesion segmentation based on preprocessing, thresholding and neural networks

J.M. Gutiérrez-Arriola, M. Gómez-Álvarez, V. Osma-Ruiz, N. Sáenz-Lechón, R. Fraile
GAMMA Group on Acoustics and MultiMedia Applications
CITSEM Research Center on Software Technologies and Multimedia Systems for Sustainability
Universidad Politécnica de Madrid
Contact: juana.gutierrez.arriola@upm.es

## Abstract

This abstract describes the segmentation system used to participate in the challenge ISIC 2017: Skin Lesion Analysis Towards Melanoma Detection. Several preprocessing techniques have been tested for three color representations (RGB, YCbCr and HSV) of 392 images. Results have been used to choose the better preprocessing for each channel. In each case a neural network is trained to predict the Jaccard Index based on object characteristics. The system includes black frames and reference circle detection algorithms but no special treatment is done for hair removal. Segmentation is performed in two steps first the best channel to be segmented is chosen by selecting the best neural network output. If this output does not predict a Jaccard Index over 0.5 a more aggressive preprocessing is performed using open and close morphological operations and the segmentation of the channel that obtains the best output from the neural networks is selected as the lesion.

## Image formats and preprocessing techniques

Three image formats have been used: RGB, YCbCr and HSV. Each channel was treated separately. Preprocessing consisted on one or several of these techniques:

- Anisotropic diffusion
- Contrast enhancement
- Color consistency
- Gamma compensation
- Color normalization

155 tests were performed on 392 images extracted from ISIC 2016 database. These images didn't present hairs, reference circles or black frames. Jaccard Index was obtained and analyzed to choose the better thresholding segmentation for each channel. Conclusions are shown in Table 1. The decision was based on the following criteria:

- The less preprocessing the better
- The highest number of Jaccard Index above 0.8
- The best mean Jaccard Index of the segmentation of the 392 images
- Number of best segmentations

Two different preprocessing techniques were applied to the Blue channel because one of them was the segmentation that presented the highest number of best segmentations among all the experiments and the other had the best mean value for the experiments carried out on the blue channel.

An Otsu thresholding was applied to each channel and the object that was closer to the center of the image was selected as the lesion. This selection was then compared with the ground truth and the Jaccard Index was calculated.

As an example, some results are shown in Figure 1 and Figure 2.

Table 1. Preprocessing applied to the different color channels

| Channel | Preprocessing with the best results in terms of Jaccard Index |
|---|---|
| R | Color normalization and contrast enhancement |
| G | Gamma compensation, color consistency, conversion to HSV, anisotropic diffusion, contrast enhancement, conversion to RGB |
| B | Contrast enhancement |
| Y | Gamma compensation, color consistency, conversion to HSV, anisotropic diffusion, contrast enhancement, conversion to RGB, conversion to gray scale |
| Cb | Anisotropic diffusion |
| Cr | Anisotropic diffusion |
| H | Anisotropic diffusion and contrast enhancement |
| S | Gamma compensation, color consistency, conversion to HSV, contrast enhancement |
| V | Gamma compensation, color consistency, conversion to HSV |
| B1 | Gamma compensation, color consistency, conversion to HSV, anisotropic diffusion, contrast enhancement, conversion to RGB |

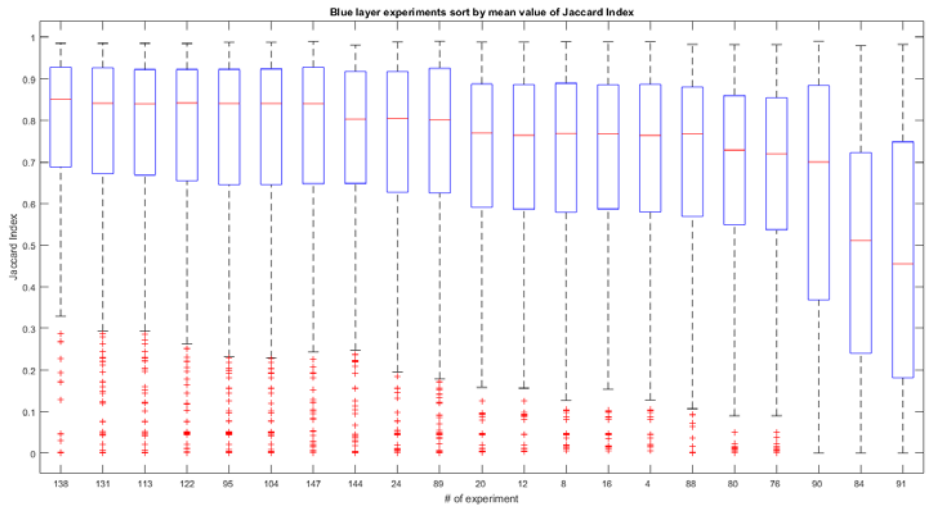

Figure 1. Results of experiments carried out on the blue channel applying different preprocessing techniques before thresholding. The experiments are sort from higher to lower mean Jaccard Index. Experiment 138 uses gamma compensation, color consistency, anisotropic diffusion and contrast enhancement.

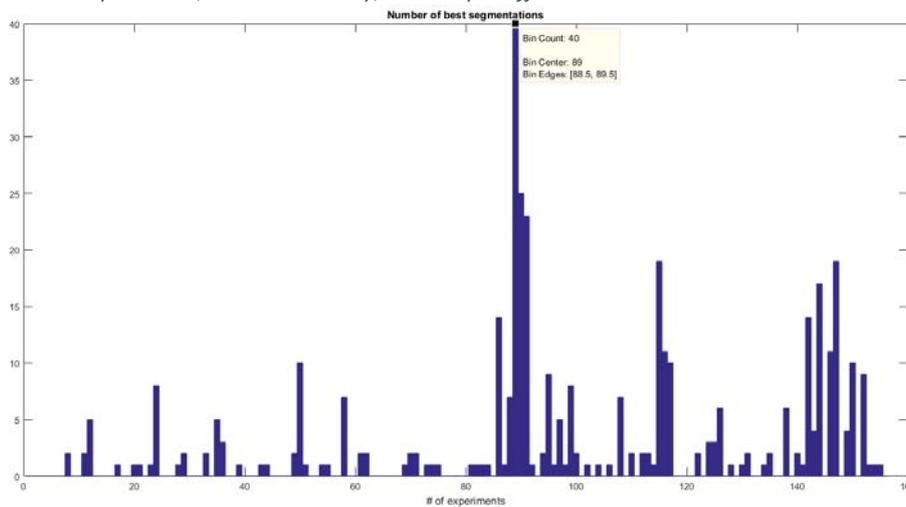

Figure 2. Number of best segmentations in each experiment. Experiment 89 uses contrast enhancement.

## Methods

The ten selected channels were segmented for the 2000 images of the training set of ISIC 2017 challenge and the following properties were obtained for the object chosen as lesion:

1. Extent
2. Solidity
3. Area/perimeter
4. Eccentricity
5. Distance from the centroid of the object to the center of the image
6. Normalized mean value of the pixels belonging to the lesion
7. Normalized mean value of the pixels of the background
8. Normalized mean value of the pixels of the Y channel belonging to the lesion
9. Normalized mean value of the pixels of the Y channel of the background
10. Normalized mean value of the pixels of the Cb channel belonging to the lesion
11. Normalized mean value of the pixels of the Cb channel of the background
12. Normalized mean value of the pixels of the Cr channel belonging to the lesion
13. Normalized mean value of the pixels of the Cr channel of the background

These parameters were the inputs of a neural network trained to predict the Jaccard Index of the segmentation. One neural network was trained for each channel with MATLAB Neural Net Fitting App choosing 20 neurons in the hidden layer.

The final segmentation system was developed in MATLAB and had the following steps:

Step 1. Resize image to 500x500 pixels.
Step 2. YCbCr and HSV images were obtained from RGB image.
Step 3. Detect black lateral bands.
Step 4. Detect reference circles.
Step 5. Detect black frames.
Step 6. Preprocess each channel as stated in Table 1.
Step 7. Otsu Thresholding. Objects smaller than the 0.2% of the image size, with extent smaller than 0.2 and solidity less than 0.4 were eliminated as candidates to be the lesion. Of the remaining objects the one whose centroid was closer to the image center was selected as the segmented lesion.
Step 8_1. Computation of inputs and output of the neural networks. If the highest output was over 0.5 (this would be the predicted Jaccard Index) it was selected as the final segmentation.
Step 8_2. If the highest output was less than 0.5 then a new thresholding was performed with the addition of an open and close morphological operation. New inputs were obtained for the neural networks and the segmentation of the channel that predicts the best Jaccard Index was selected as lesion.
Step 9. Resize the image and the segmentation mask to the original size.

## Results and future work

The system was tested with the validation set of the ISIC 2017 challenge containing 150 images. Average Jaccard Index was 0.705. 120 images were segmented in step 8_1 with an average Jaccard Index of 0.787 and 29 were segmented in step 8_2 with an average Jaccard Index of 0.361.  In Figure 3 and Figure 4 a successful and a failed segmentation are shown.
In future works we will try to obtained a better segmentation of the images that don't have a good Jaccard Index prediction. We have tried with an algorithm based on watershed transform and, although we have promising results, the system is not yet completely automatic.

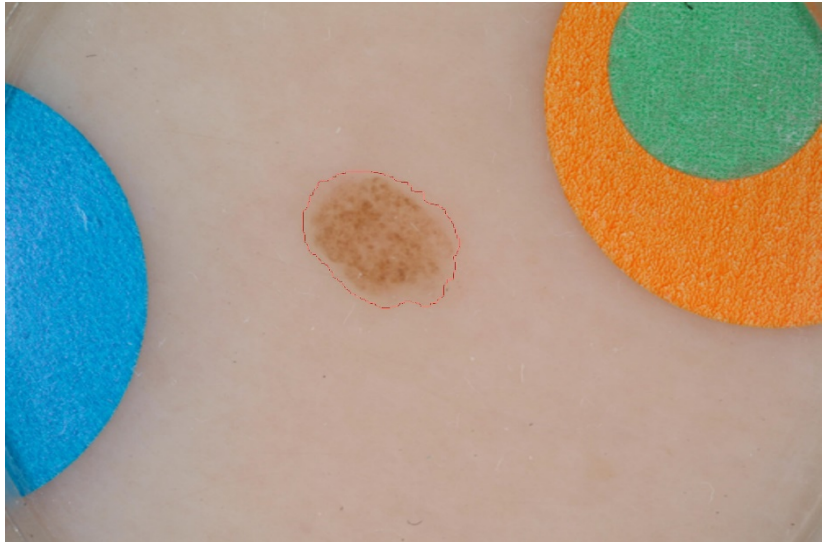

*Figure 3. Successful segmentation, lesion is outlined.*

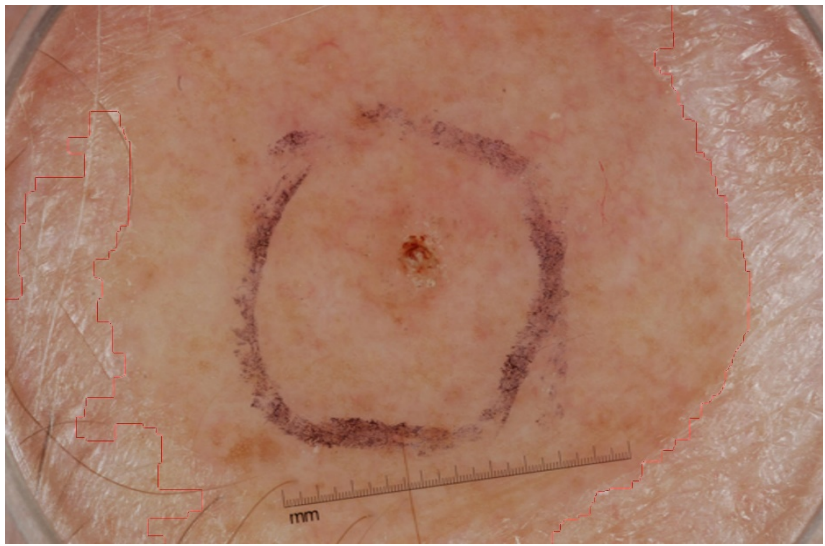

*Figure 4. Failed segmentation, detected lesion is almost all the image.*